\documentclass[letterpaper, 10 pt, conference]{ieeeconf}  

\IEEEoverridecommandlockouts                              
\usepackage{times} 
\usepackage{amsmath} 
\usepackage{amssymb}  
\usepackage{float}
\usepackage{wrapfig,lipsum,booktabs}
\usepackage{siunitx}
\usepackage[pdftex]{graphicx}

\usepackage[OT1]{fontenc}

\usepackage[pdfusetitle]{hyperref}

\usepackage{cite}
\usepackage[capitalise]{cleveref}
\usepackage{subcaption}

\usepackage{mathtools}

\usepackage{microtype}
\usepackage[font=small]{caption}

\title{\LARGE \bf
Low-Cost Fiducial-based 6-Axis Force-Torque Sensor
}

\author{Rui Ouyang$^{1}$, Robert Howe$^{2}$
\thanks{$^{1,2}$School of Engineering and Applied Sciences, Harvard University, Cambridge, MA.
{$^{1}$\tt\small nouyang@g.harvard.edu}}
}

\begin{document}

\maketitle
\thispagestyle{empty}
\pagestyle{empty}

\begin{abstract}

    Commercial six-axis force-torque sensors suffer from being some combination
    of expensive, fragile, and hard-to-use. We propose a new fiducial-based
    design which addresses all three points. The sensor uses an inexpensive
    webcam and can be fabricated using a consumer-grade 3D printer. Open-source
    software is used to estimate the 3D pose of the fiducials on the sensor,
    which is then used to calculate the applied force-torque.  A browser-based
    (installation free) interface demonstrates ease-of-use. The sensor is very
    light and can be dropped or thrown with little concern. We characterize our
    prototype in dynamic conditions under compound loading, finding a mean $R^2$
    of over 0.99 for the $F_x, F_y, M_x$, and $M_y$ axes, and over 0.87 and 0.90
    for the $F_z$ and $M_z$ axes respectively. The open source design files
    allow the sensor to be adapted for diverse applications ranging from robot
    fingers to human-computer interfaces, while the sdesign principle
    allows for quick changes with minimal technical expertise. This approach
    promises to bring six-axis force-torque sensing to new applications where
    the precision, cost, and fragility of traditional strain-gauge based sensors
    are not appropriate. The open-source sensor design can be viewed at
    \url{http://sites.google.com/view/fiducialforcesensor}.

\end{abstract} 

\section{Introduction}

\subsection{Motivation}

    Force-torque sensors are used extensively in both industry and research.
    We focus here on the use of these sensors in two examples: robotic grasping, where they
    are used to provide tactile feedback (e.g. detecting when contact is made),
    and in human computer interaction.
    However, commercial six-axis force-torque sensors can be both expensive
    and fragile. This combination makes them tricky to use for grasping, where
    controlled contact is desired, but a small coding error could easily smash
    and overload the sensor. One of the most common types of sensors, the
    ATI force/torque sensor, costs tens of thousands of dollars and
    relies on strain gauges that are fragile and have to be surrounded in a
    bulky package.
    For these reasons, we are motivated to consider new sensor designs that could
    promote the use of tactile data in the robotics community through being a
    combination of cheaper, easier to use, and more robust. 

    \begin{figure}[t]
        \centering
            \includegraphics[width=\linewidth]{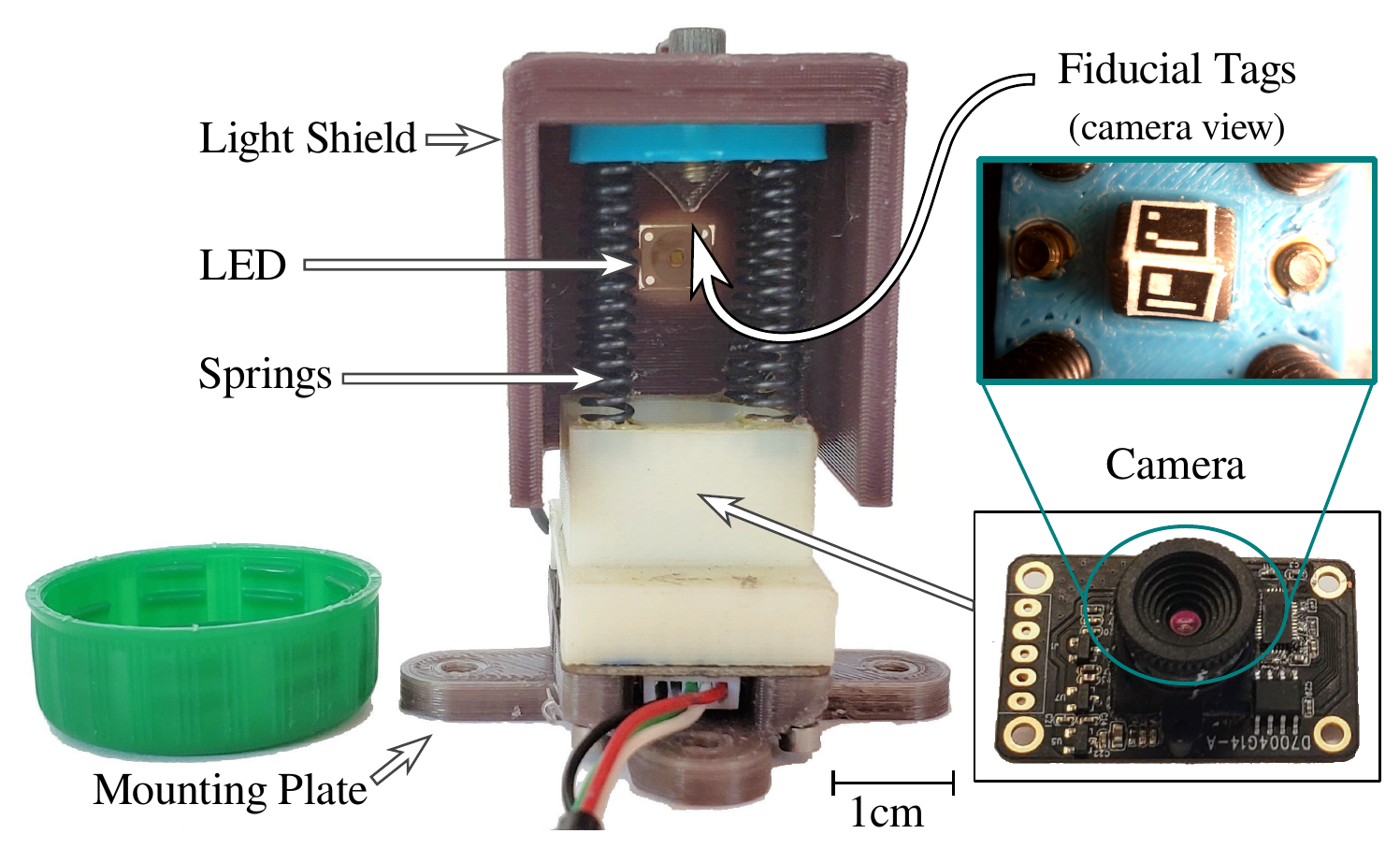}
            \caption{Consumer webcams and a printed fiducial markers can be used to create a
            six-axis force-torque sensor. We used four springs to build a
            platform free to move in all angular directions. We affixed two
            printed fiducials to the platform, and then aimed a consumer
            camera up at them. To the right, the camera view 
            reveals the tag location. The tags are glued to the light
            shield, which is removable, allowing for easy design changes.
            Note that cardstock, which was removed for picture clarity, was used
            to diffuse the LED and avoid overexposing the camera. Green bottle cap is for scale.}
            \label{fig:introlabelled}
    \end{figure}

\subsection{Related Work}

    Multiple designs have emerged recently taking advantage of the rich
    information available from consumer webcams. Even low-end webcams will
    output 640x480 RGB images at 15 frames-per-second (fps). The webcam-based
    sensors are particularly easy to manufacture and wire. Notable examples
    include the Gelsight \cite{Yuan2015MeasurementOS}, GelForce
    \cite{Sato2010FingerShapedGS}, TacTip \cite{WardCherrier2018TheTF}, the
    Fingervision \cite{Yamaguchi2016CombiningFV}, and others. These sensors rely
    on cameras facing markers embedded in transparent or semi-transparent
    elastomer (often with supplemental LED lighting). These can be used to
    estimate shear, slip, and force, but tend not to do well in cases where the
    object hits the side of the finger instead of dead on. They also require
    casting elastomers.

    Several MEMS multi-axis force-torque sensors have been developed, which use
    the same principle of creating a device free to deflect into multiple axes,
    but then measures them using capacitative \cite{Beyeler2009ASM} or
    piezoresistive \cite{estevez20126} means. In
    \cite{Cappelleri2009TwodimensionalV} the deflection is measured using a
    camera as well, a CCD camera mounted to a microscope, however the device
    only measures two directions of force.

    Prior work used MEMS barometers to create six-axis force-torque sensors
    with very low parts cost and good durability \cite{Guggenheim2017RobustAI}.
    However, fabricating the sensor requires specialized lab equipment such as a
    degassing machine.

    Other work explored estimating fingertip force via video, but only for human
    fingers \cite{yu2008, Sartison2018FingerGF}. Commercial sensors like
    the Spacemouse and the OptoForce use similar ideas, but rely on custom
    circuitboards for a ranging sensor inside. In contrast, our work is
    straightforward to fabricate even for users unfamiliar with electronics. 

\subsection{Contributions}

    In this paper, we investigate novel combinations of readily-accessible
    technologies to create six-axis force-torque sensors that are inexpensive,
    require minimal expertise to design and build, and are easily customized for
    diverse applications.
    
    The proposed novel type of sensor makes six-axis force-torque measurements by
    tracking position and orientation displacement using the 3D pose estimate
    from fiducial tags, and uses a linear fit between displacement and applied
    force-torque.  Fiducials are markers used to help locate objects or serve as
    points of reference. They can be found in robotics and augmented reality
    applications, where they usually take the form of printed paper markers
    glued onto various objects of interest. Sensors employing these fiducials
    operate by detecting the sharp gradients that are created between black and
    white pixels, such as
    one might find on a checkerboard. An example of two fiducials can be found
    in the top right of the labelled diagram of our sensor at
    \cref{fig:introlabelled}. Using the known geometry of the tag (e.g.
    perpendicular sides of checkeboard), as well as known tag size and
    pre-determined camera calibration matrix, the 3D object pose (location and
    orientation) of the object can be estimated. This calculation is known as
    the solving the Perspective-\textit{n}-Point (PnP) problem.  We created prototypes
    utilizing two open-source tag protocols, AprilTags \cite{Olson2011AprilTagAR} and ArUco markers
    \cite{GarridoJurado2014AutomaticGA}; pictured in \cref{fig:introlabelled}
    are two ArUCo markers. 
    
    In the following sections, we begin with the design and fabrication process
    for our sensor. We follow with a theoretical analysis of how the
    sensor design parameters affect resolution, sensitivity, measurement range,
    and bandwidth. We also present an analysis of data collected from a
    prototype sensor. We conclude with a discussion of the advantages and
    limitations of this sensor.     

\section{Design}

    \subsection{Sensor Design}

    \begin{figure}[t]
        \includegraphics[width=\linewidth]{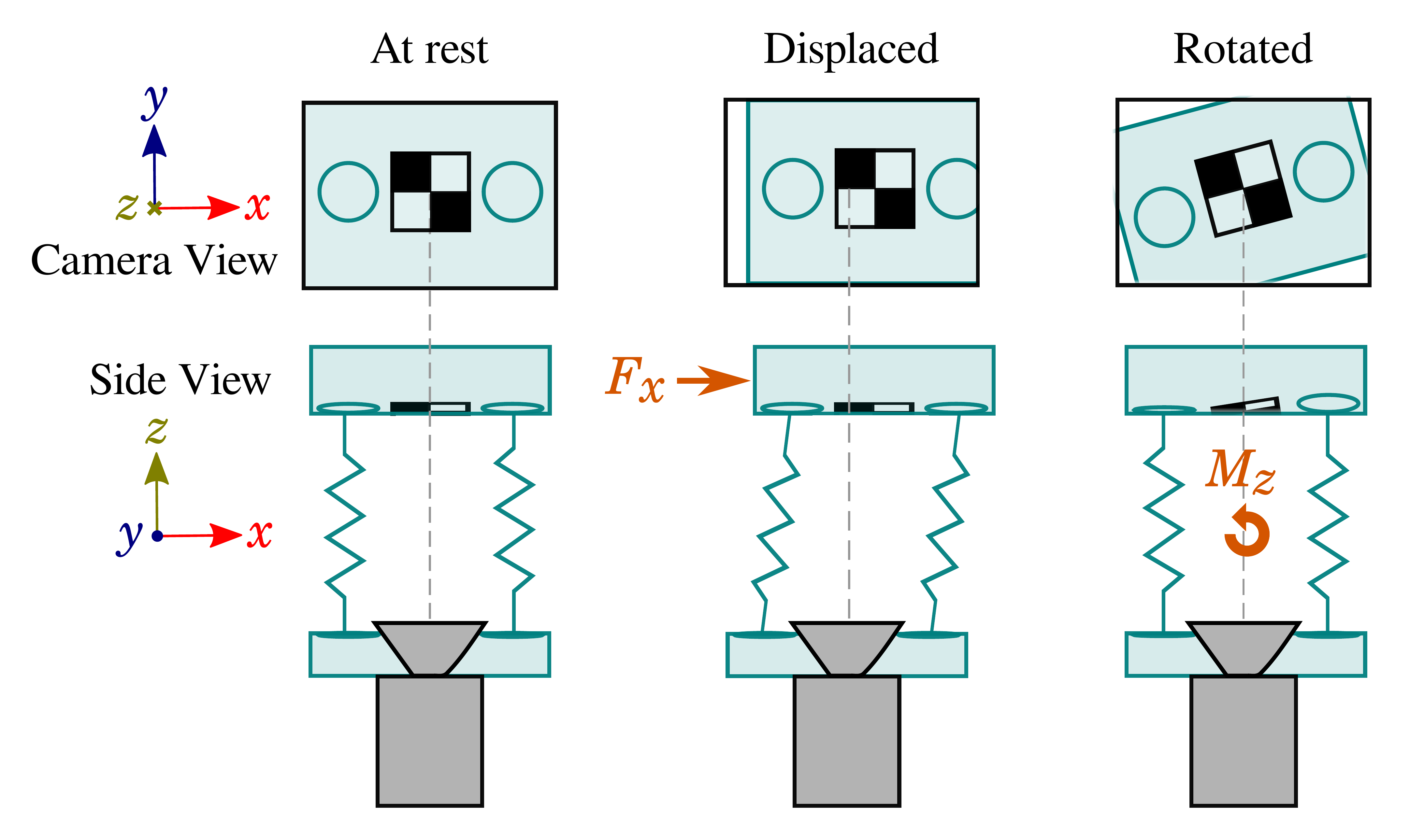}
        \caption{The top row shows the viewpoint from the camera when
        different forces or torques are applied (the dotted grey line shows the
        center of the camera view). By tracking the movement of the fiducial(s), we can
        derive the force and torque exerted on the sensor.}
        \label{fig:diagram}
    \end{figure}

    At a high level, the sensor consists of two main parts: a base and a
    platform above the base. The platform is connected to the base with 4 springs and can
    move in all directions with respect to the base. Two fiducial tags were
    glued to the underside of the platform. Then, a webcam pointed up at the
    tags was installed at the base. As force or torque is applied to the
    platform, the tags translate and rotate accordingly. The camera is used to 
    track the 3D pose of the tags. Should there be a suitably linear
    relationship between the displacement and the force-torque applied, 
    a short calibration procedure using known weights can be used to collect
    datapoints for regression. Given a known linear fit, the sensor can then output force and
    torque measurements. \cref{fig:diagram} shows the principle
    behind this fiducial-based force sensor.

    \subsection{Design Goals}
    \label{sec:designgoals}

    When designing the sensor prototype, a few considerations were made. First
    and foremost, the sensor needs to be sensitive to all six degrees of freedom
    (displacement in $x$, $y$, $z$ and rotation in yaw, pitch, roll). For
    illustrative purposes, the following analysis is performed in terms of
    specific specification values that are appropriate for a sample robot
    gripper. Alternate values for other use cases such as human-computer
    interfaces can be easily substituted. For grasping, between $\pm$ 40~N is
    realistic, and sensitivity of at least 1/10 N is desirable. Qualitatively,
    we want the sensor to be small (for grasping applications, the sensor should
    be roughly finger-sized), inexpensive, and robust. The sensor should allow for
    rapid prototyping and easy customization with minimal technical expertise.
    The sensor should be not only easy to fabricate, but also easy to use.

    \subsection{Fabrication}

    \begin{figure}
        \centering
        \begin{subfigure}[t]{0.37\linewidth}
            \centering
            \includegraphics[width=\linewidth]{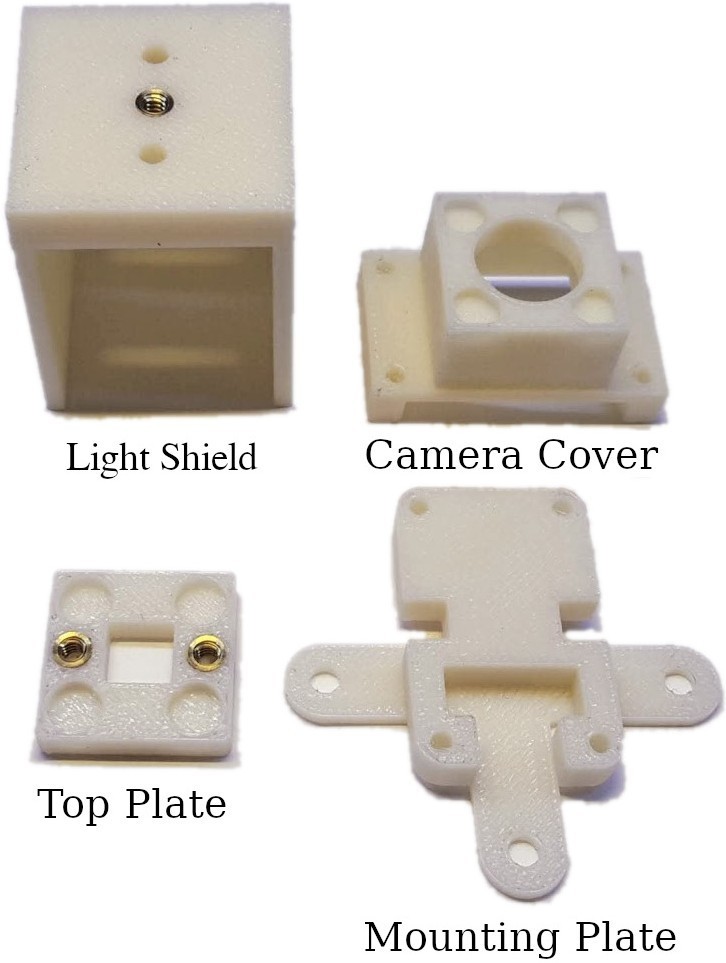}
        \end{subfigure}
        \hfill
        \begin{subfigure}[t]{0.58\linewidth}
            \centering
            \includegraphics[width=\linewidth]{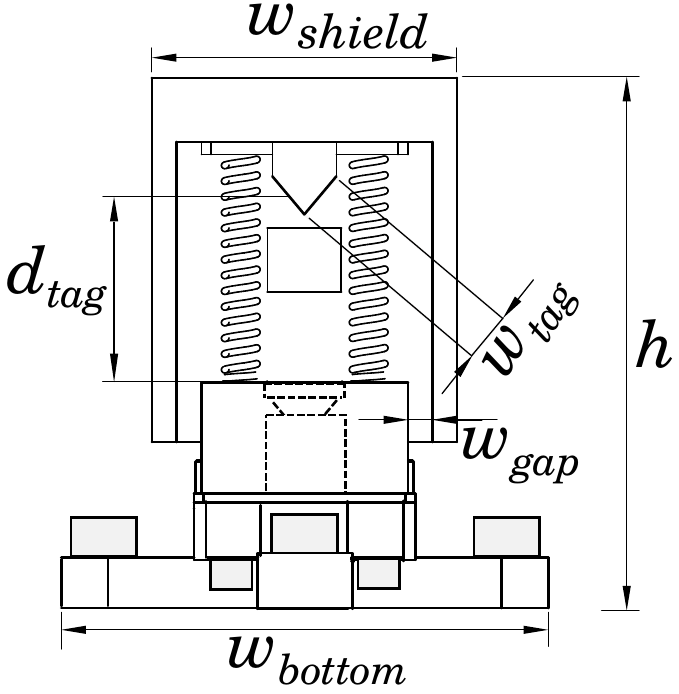}
            \label{fig:sensorsolidworks}
        \end{subfigure}
        \caption{Left, the four 3D printed parts are shown. Right, 
            a diagram of the sensor as mounted to the commercial force sensor
            (in grey on the bottom) used in our experiments. The footprint
            of the sensor itself is the same as the camera circuitboard, 35.7~mm by
            22.5~mm. The sensor height $h$ = 51~mm, while the camera lens is
            approximately $d_{tag}$ = 21~mm from the center of the tags. The
            light shield is offset on all three sides by $w_{gap}$ = 2.5~mm gap
            from the camera cover, and has width $w_{shield}$ = 31~mm.
            The fiducials are each
            $w_{tag}$ = 3.8~mm wide (or $4.5$~mm including the white border).
            The mounting plate attaches to mounting holes in the force sensor
            and has width $w_{bottom}$ = 45~mm.}
        \label{fig:pieces}
    \end{figure}

    \subsubsection{Physical Fabrication}

    The four pieces in \cref{fig:pieces} (figure includes dimensions) are
    3D-printed in two to three hours on an inexpensive consumer-grade device
    (Select Mini V2, Monoprice). Epoxy
    is used to glue the springs into the camera cover and top plate. The tags
    are printed on paper and glued in. A small piece of white cardstock is used
    to diffuse the LED (in the future, this would be built into the 3D design).
    Conveniently, the pose estimate is
    relative to the camera frame, and the sensor relies only on relative
    measurements, so the tag placement can be imprecise. The LED is mounted 
    in and connected to a 3.3~V power source. The heat-set thread inserts (for bolting the
    light shield to the platform) are melted in with a soldering iron. The
    camera is placed between the mounting plate and camera
    cover and then everything is bolted together. The springs are steel
    compression springs available online as part of
    an assortment pack from Swordfish Tools. The spring dimensions are
    2.54~cm long, 0.475~cm wide, and wire width of 0.071~cm,
    with a stiffness of approximately $\SI{0.7}{N/mm}$.  Fabrication can be
    completed in a day. The actual assembly, given a complete set of hardware
    and tools, can be completed in 30 minutes, depending on the epoxy setting
    time.

    \subsubsection{Usage and Software}

    The only data cable used is the USB from the webcam to the computer.  
    On the computer, the OpenCV Python library \cite{opencv_library} (version
    4.1.2) is used to
    detect the ArUco markers in the video feed. We used a commercial
    force-torque sensor to characterize our sensor, for which we used another
    freely available Python library (see \cite{python_optoforce}). The data from
    the commercial sensor (Model HEX-58-RE-400N, OptoForce, Budapest, Hungary)
    and the markers are read in parallel threads and timestamped, then recorded
    to CSV. Python is used for further analysis.

\begin{figure}[htbp]
        \centering
        \includegraphics[width=0.9\linewidth]{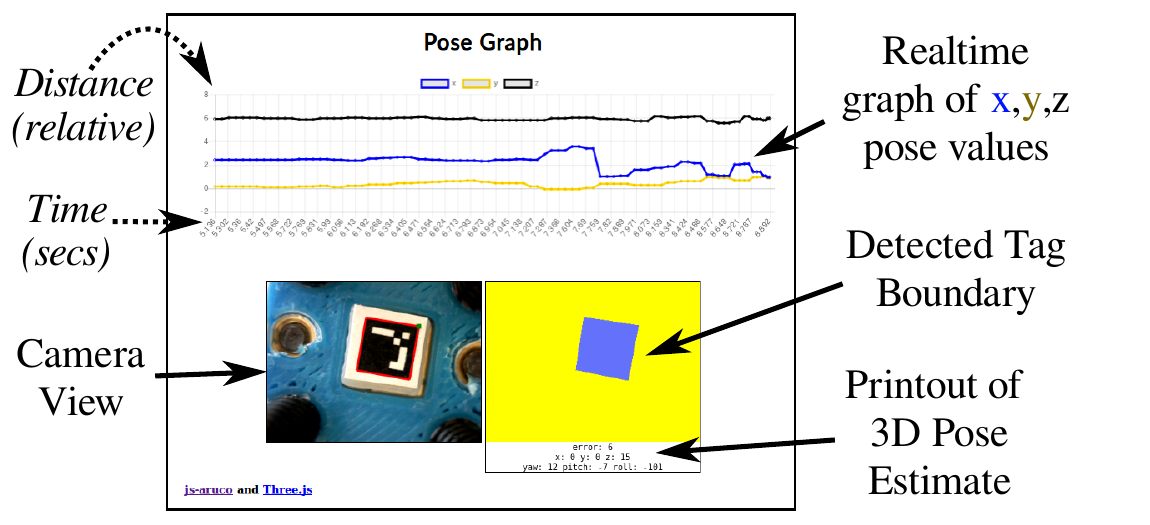}
        \caption{Our prototype JavaScript-based interface (modified from
            the \textit{Js-aruco} library example) \cite{JS-aruco}. In this way,
            sensor data can be read just by loading a webpage.}
        \label{fig:JS_GUI}
    \end{figure}

    By using a consumer webcam, sensor reading is also possible
    without installing Python. To demonstrate this, we developed a simple
    interface using a Javascript ArUco tag detector library (see
    \cite{JS-aruco}). \cref{fig:JS_GUI} shows a graphical user interface (GUI)
    that plots the $x$, $y$, and $z$-axes of the 3D pose estimate for a single
    tag.

    In theory, the sensor reading can be done on-the-go with a smartphone and a
    wireless or USB-C webcam (such as inexpensive endoscope inspection cameras found
    online).

    \subsubsection{Calibration}

    Although we calibrated using a commercial force-torque sensor, the same can
    be achieved with a set of weights and careful clamping. The sensor can be
    clamped sideways to a sturdy surface to calibrate the $x$- and $y$-axes. A set of
    known weights is then attached to the center bolts on the light shield piece
    via a string. The same procedure can be applied to calibrate the $z$-axis,
    with the sensor clamping upside down to a tabletop. Finally, weights can be
    applied to the two side bolts to produce known torques while hanging upside
    down or sideways.

\section{Analysis}

    Considering the above design goals, there are a few primary concerns
    amenable to theoretical analysis: the sensor resolution, sensitivity,
    force range, and bandwidth. Here, sensor resolution is defined in bits
    (relative terms) and sensitivity in millimeters and degrees. 

    \subsection{Resolution}

    \begin{figure}
        \centering
        \begin{subfigure}[t]{0.40\linewidth}
            \centering
            \includegraphics[height=3cm]{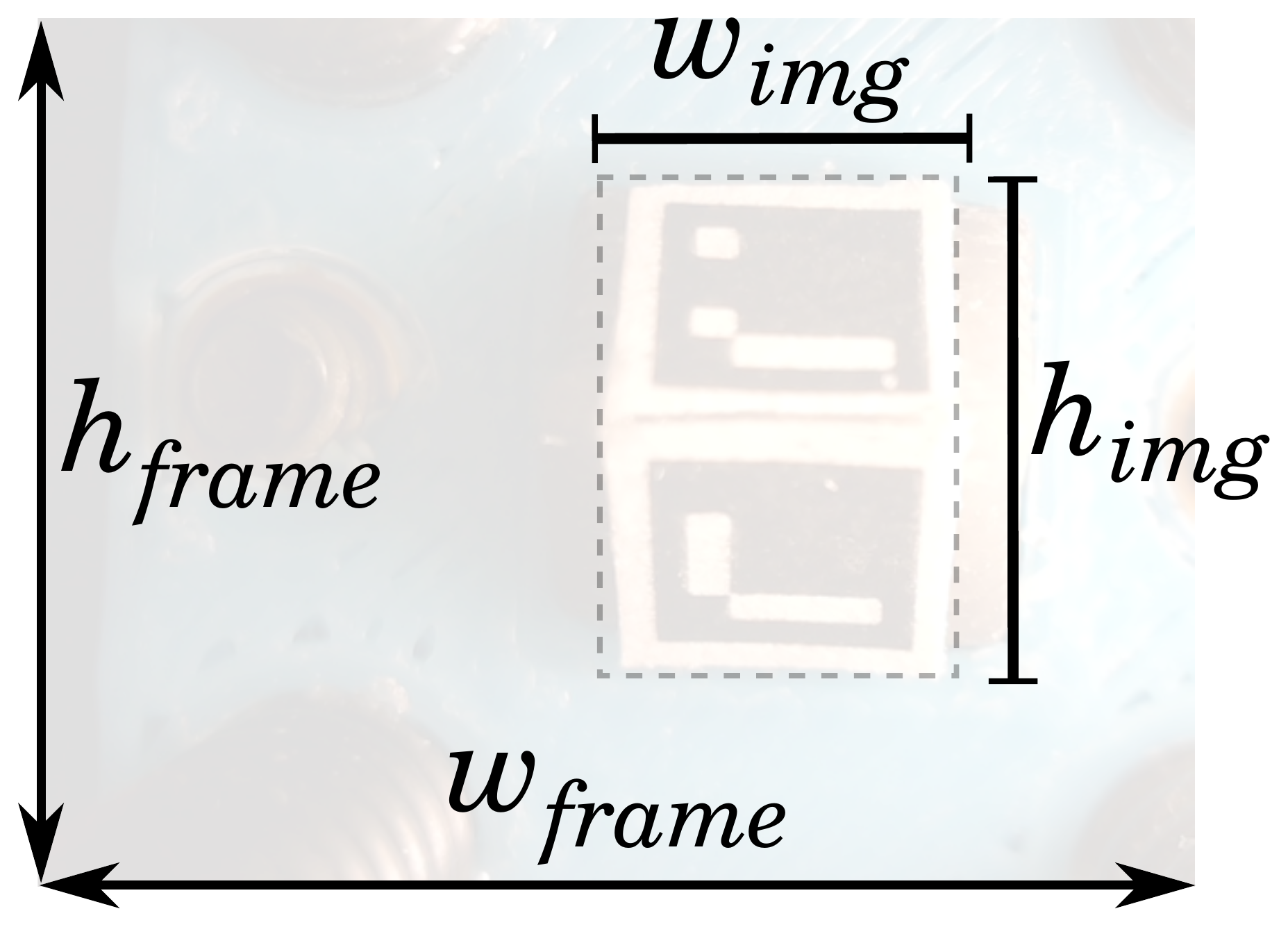}
            \caption{A frame from the camera. $w_{frame} \times h_{frame} = 640
                \times 480$~pixels, and $w_{img} \times h_{img} = 150 \times 240$~pixels. }
            \label{fig:xy_resolution}
        \end{subfigure}
        \hfill
        \begin{subfigure}[t]{0.55\linewidth}
            \centering
            \includegraphics[height=4cm]{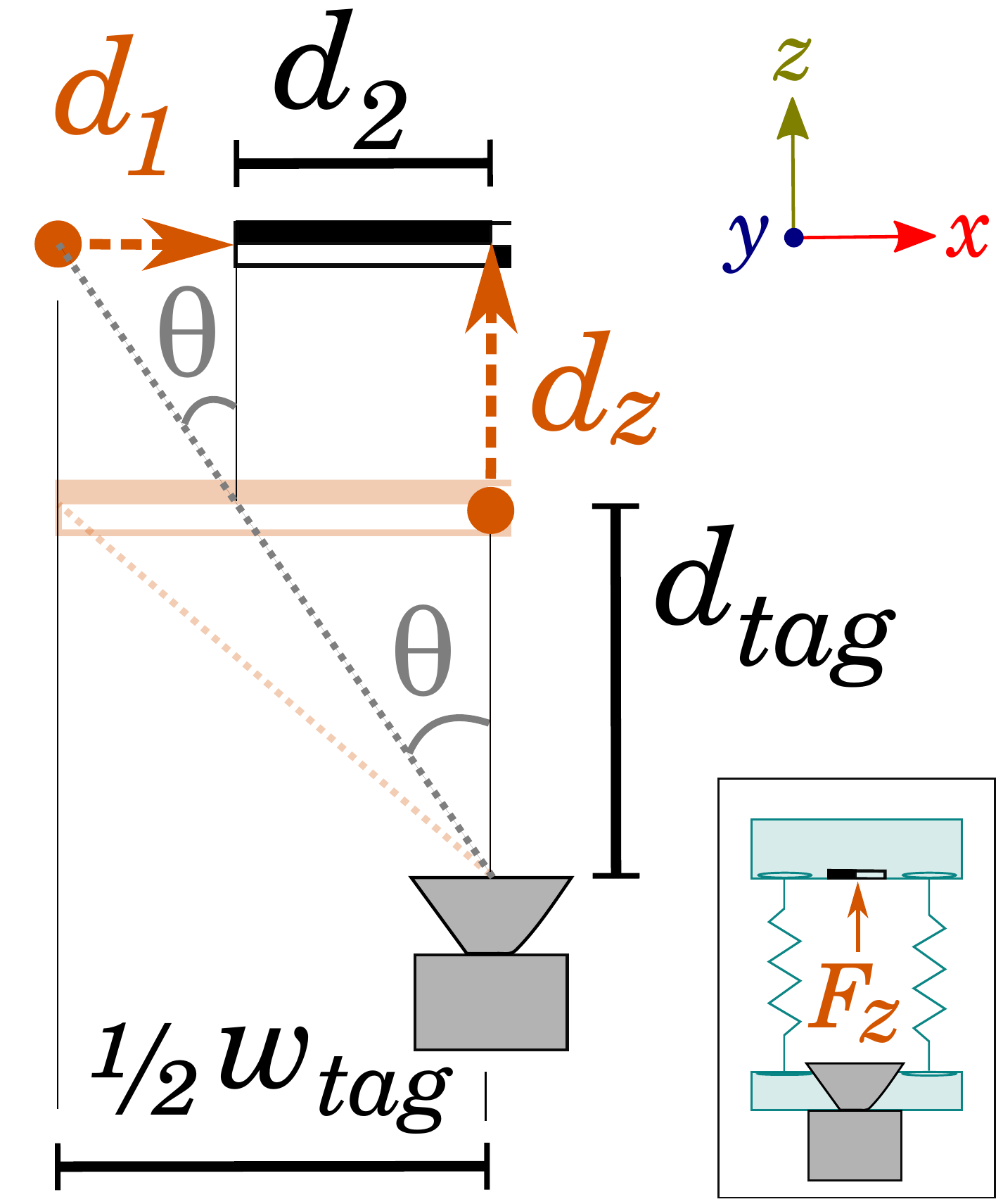}
            \caption{Side view. As per \cref{fig:pieces}, $d_{tag}=21$~mm and
                $w_{tag}= 4.5$~mm. Light orange indicates original tag position
            before displacement. Inset shows force applied.}
            \label{fig:zres}
        \end{subfigure}
        \caption{Sensitivity calculation diagrams.}
    \end{figure}

    Let us conservatively estimate the discernible resolution of the tag
    system to be $d_R = 1/4$ pixel, or $C= 4$ counts per pixel. This factor exists because we
    have more than just binary information (1 bit) for every pixel. For
    instance, if a black/white intersection is halfway between two pixels, the
    pixels will be gray. (Tag algorithms also use the known grid geometry to
    achieve subpixel resolution -- see the \textit{cornerSubPix} function in the OpenCV
    library).

    In that case, we can determine the resolution of the sensor itself
    geometrically, by looking at the number of pixels. The fact that the tags must
    stay on-screen limits the sensor resolution. 
    
    We can characterize an approximate $y$-axis resolution $r_y$ of the camera by taking the
    number of pixels available, multiplying by $C$, and converting our counts into bits.
    \begin{align}
        r_y &= \lfloor \log_2 \left( C \cdot (h_{frame} - h_{img}) \right) \rfloor + 1
    \end{align}

    For instance, the calculations for our sensor prototype are as follows. In the $y$-axis,
    \begin{align}
        r_y &= \lfloor \log_2{\left(4 \cdot (480 - 240)\right)} \rfloor \\
        r_y &= 10 ~\text{bits}
    \end{align}

    In the $x$-axis, repeating the same calculations we have    
    \begin{align}
        r_x &= \lfloor \log_2 \left( C \cdot (w_{frame} - w_{img}) \right) \rfloor + 1 \\
        r_x &= \lfloor \log_2{\left(4 \cdot (640- 150)\right)} \rfloor + 1 \\
        r_x &= 11 ~\text{bits}
    \end{align}

    In the $z$-axis, our limitation is the same as the $y$-axis, so we have $r_z = 11 ~\text{bits}$.

    \subsection{Sensitivity}

    Let us now calculate the sensitivity of the sensor. We will start by
    looking at the minimum detectable travel in each of the $x$, $y$, and $z$-axes.

    \subsubsection{Translational Sensitivity}

    In the $x$ and $y$ directions, we can measure the mm/px at rest (the sensor
    resolution varies a bit since the tag gets larger or smaller depending on
    the $z$ distance). Roughly, the tag measures $\SI{4.5}{mm}$ and appears as
    $w_{tag} = 150$~pixels in the image. Assuming as above that we can discern 4
    counts per pixel, the theoretical sensitivity is
    \begin{align}
        s_y &= \frac{h_{frame} ~\SI{}{(mm)}}{h_{frame} ~\SI{}{(px)}} d_R 
            = \frac{4.5}{150} \cdot \frac{1}{4} 
            = \SI{0.0075}{mm} 
    \end{align}

    For the $z$-axis sensitivity, we consider that the tag will get smaller as
    it displaces in the $+z$ direction. Using a simple geometrical model (see
    \cref{fig:zres}), given that the smallest detectable change in $xy$ plane is
    $1/4$~pixel, we can calculate what is the resulting change in $z$. 

    Using similar triangles, we see that 
    \begin{align}
        \frac{d_1}{d_z} &= \frac{d_2}{d_{tag}} \\
        d_1 + d_2 &= w_{img}/2 \\
        d_2 &= (w_{img}/2) - d_1
    \end{align}

    We would like to work in $\SI{}{mm}$, therefore we use the fact that the tag is
    4.5~mm and appears as 150~px.
    \begin{align}
        d_1 = d_R = 1/4 \; \text{px} \cdot \frac{\SI{4.5}{mm}}{\SI{150}{px}} 
        &= \SI{0.0075}{mm} \\
        d_2 = \frac{4.5}{2} - 0.0075 &= \SI{2.2425}{mm} \\
        s_z = d_z = \frac{d_1}{d_2} d_{tag} = \frac{0.0075}{2.2425} \cdot 21 &= \SI{0.07}{mm}
    \end{align}

    \subsubsection{Rotational Sensitivity}

    \begin{figure}
        \centering
        \begin{subfigure}[t]{0.45\linewidth}
            \centering
            \includegraphics[height=4cm]{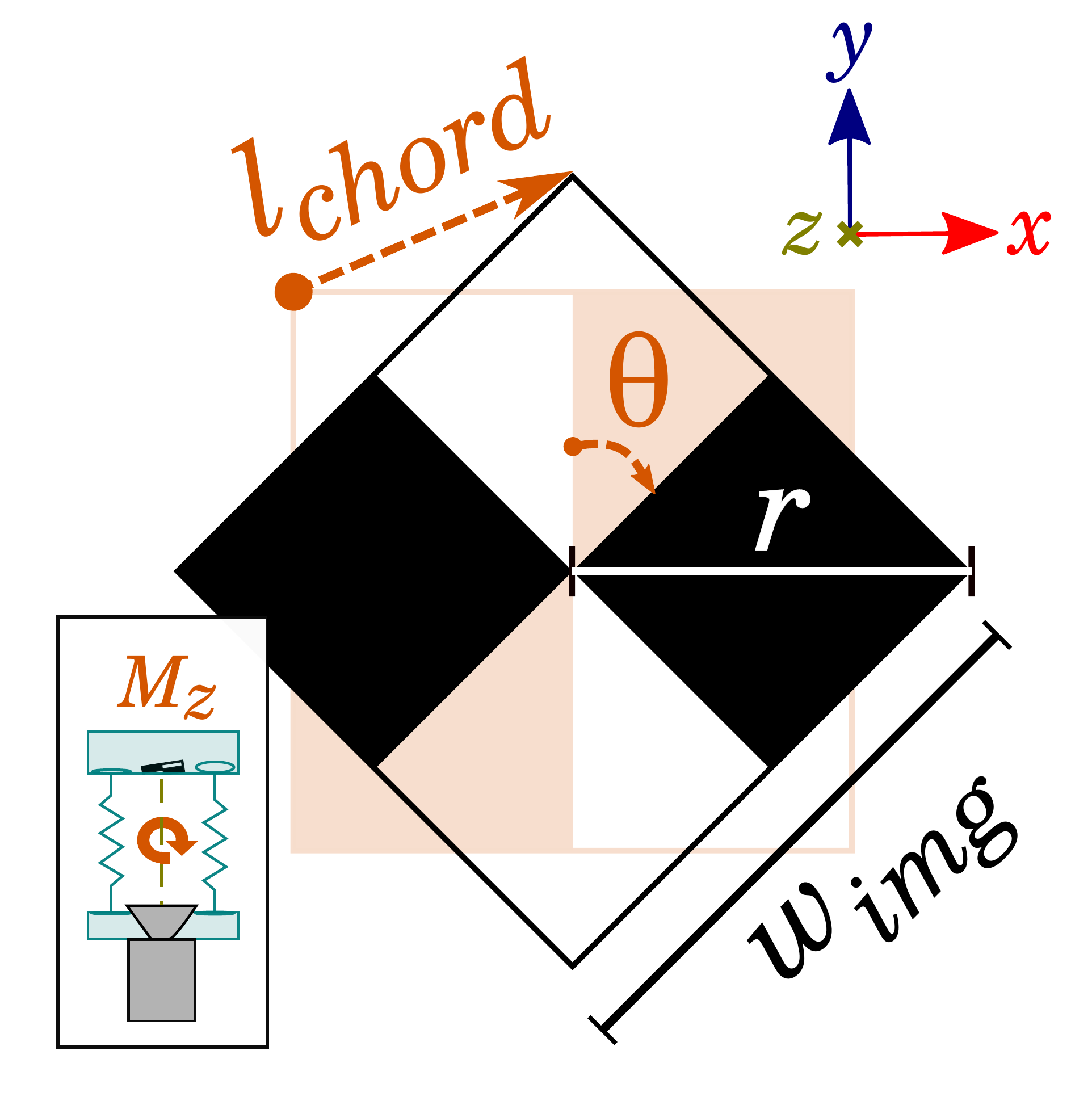}
            \caption{Camera view. The light orange shows original tag
            orientation before a \ang{45} turn around the $z$ axis.}
            \label{fig:tauz_res}
        \end{subfigure}
        \hfill
        \begin{subfigure}[t]{0.50\linewidth}
            \centering
            \includegraphics[height=3.8cm]{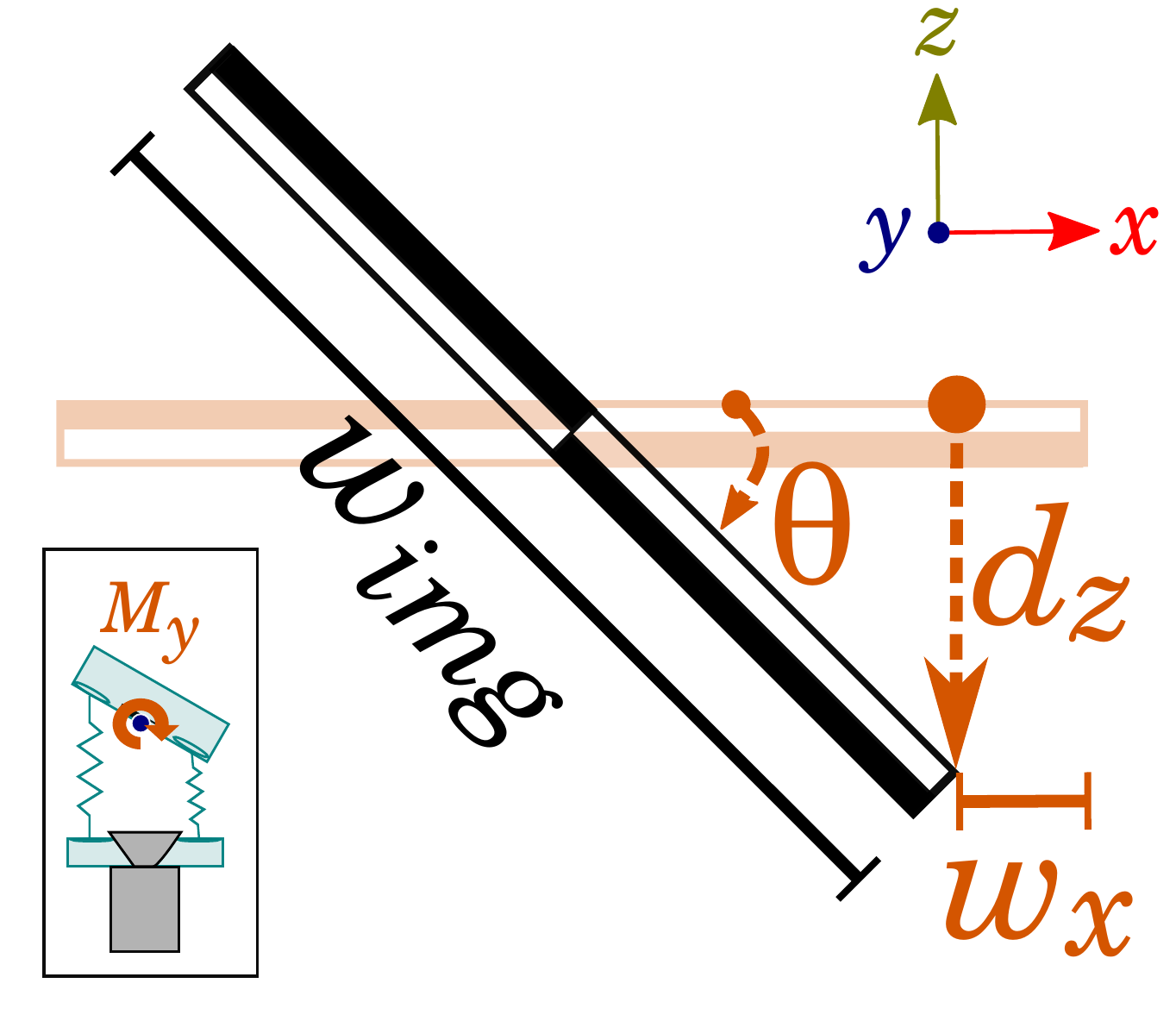}
            \caption{Top-down view. The light orange shows original tag
                orientation before a \ang{45} turn around the $y$-axis (or
                equivalently $x$-axis).}
                \label{fig:tauxy_res}
        \end{subfigure}
        \caption{Sensitivity calculation diagrams. Insets show
        applied moment.}
    \end{figure}
    
    For rotation about the $z$ axis, we can calculate the chord length in pixels
    traveled when a tag is rotated 45~degrees (about its center), and use the
    same assumption of four counts per pixel to estimate our rotational
    sensitivity. Geometrically, we know that
    \begin{align}
        l_{chord} &= 2 ~r \sin\frac{\theta}{2} 
    \end{align}

    In our case, with $w_{img} = \SI{150}{px}$, we see that
    \begin{align}
        \mathit{r} &= \sqrt{2} \cdot w_{img} /2 \\
        l_{chord} &= 2 \sqrt{2} \cdot 150/2 \cdot \sin \frac{\pi /4}{2} = \SI{81.18}{px} \\
        s_{\tau z} &= \frac{\theta}{l_{chord}} \cdot d_R 
                    = \frac{81.18}{150} \cdot \frac{1}{4} 
                    = \ang{0.14}
    \end{align}

    For rotation about the $x$ and $y$-axes, the analysis becomes a matter of
    determining the $z$-axis change in mm, and using that to determine the
    pixels changed in the $x$-$y$ plane.
    Consider a 45~degree rotation around the $z$-axis of a tag that starts out flat
    (facing the camera), as shown in \cref{fig:tauxy_res}. Using $w_{img} =
    \SI{150}{px}$ as before, the $z$ sensitivity is as follows:
    \begin{align}
        w_{img}/2 &= \sqrt{2} \cdot d_z \\
        d_z + w_x &= w_{img}/2 \\
        w_x &= 0.5 \; w_{img} - \frac{0.5 \; w_{img}}{\sqrt{2}} = \SI{21.97}{px} \\
    s_{\tau xy} &= \frac{\theta}{w_x} d_R 
    = \frac{\ang{45}}{21.97} \cdot \frac{1}{4} = \ang{0.51}
    \end{align}

    \subsection{Notes on \textit{z}-axis measurements}
    \label{sec:zres}

    Intuitively, we expect that the sensor is much less reliable in the $z$ 
    displacement direction. For movement along the $x$ and $y$-axes
    axes, the camera sees the entire set of black/white intersections moving
    left or right.  
    
    For the same reason, in the single tag setup it would be easy to
    detect rotations about the $z$-axis, and difficult to detect rotations around
    the $x$ and $y$-axes. Data collected from this initial (single-tag) design
    exactly reflected the aforementioned issue. Consequently, the design was
    enhanced with two tags oriented at 45 degrees to the camera. This proved
    sufficient for recovering all six force/torque axes.

    \subsection{Force Range Versus Sensitivity}
    \label{sec:40Ncalc}

    There is a clear trade-off between sensitivity (minimum detectable change in
    force) and the maximum force range. As an example, for a
    desired force range $F_{range} = \pm\SI{1}{N} = \SI{2}{N}$ (close to the observed force
    range for our prototype), and a maximum
    displacement of $y_{range} = h_{frame} - h_{img}$, 
    the $y$ sensitivity $s_y$ in Newtons is as follows.
    \begin{align}
        s_{y} &= \frac{F_{range}}{ C \cdot  y_{range}} 
                = \frac{2}{4 \cdot (480 - 240)} = \SI{0.0021}{N}
    \label{eqn:maxdisplacement}
    \end{align}

    Our $s_y$ is thus 2.1~mN (given our assumption of $d_R = 0.25$). 
    Similarly, for the $x$-axis we find a
    sensitivity $s_x$ = $\SI{1.0}{mN}$ at this force range.  Now consider
    instead the grasping use case, with a desired force range of $\pm$ 40~N, and
    desired sensitivity of at least 0.1~N. If we scale the calculations in
    \cref{eqn:maxdisplacement} by
    40 to get a $\pm$ 40~N force range while keeping the other parameters the
    same, the sensor has 0.04~N and 0.08~N sensitivities in the $x$ 
    and $y$ directions respectively.
    
\section{Sensor Prototype Evaluation}

    \subsection{Linearity}

\begin{figure}[t]
        \centering
        \begin{subfigure}[b]{0.45\linewidth}
            \centering
            \includegraphics[width=\linewidth]{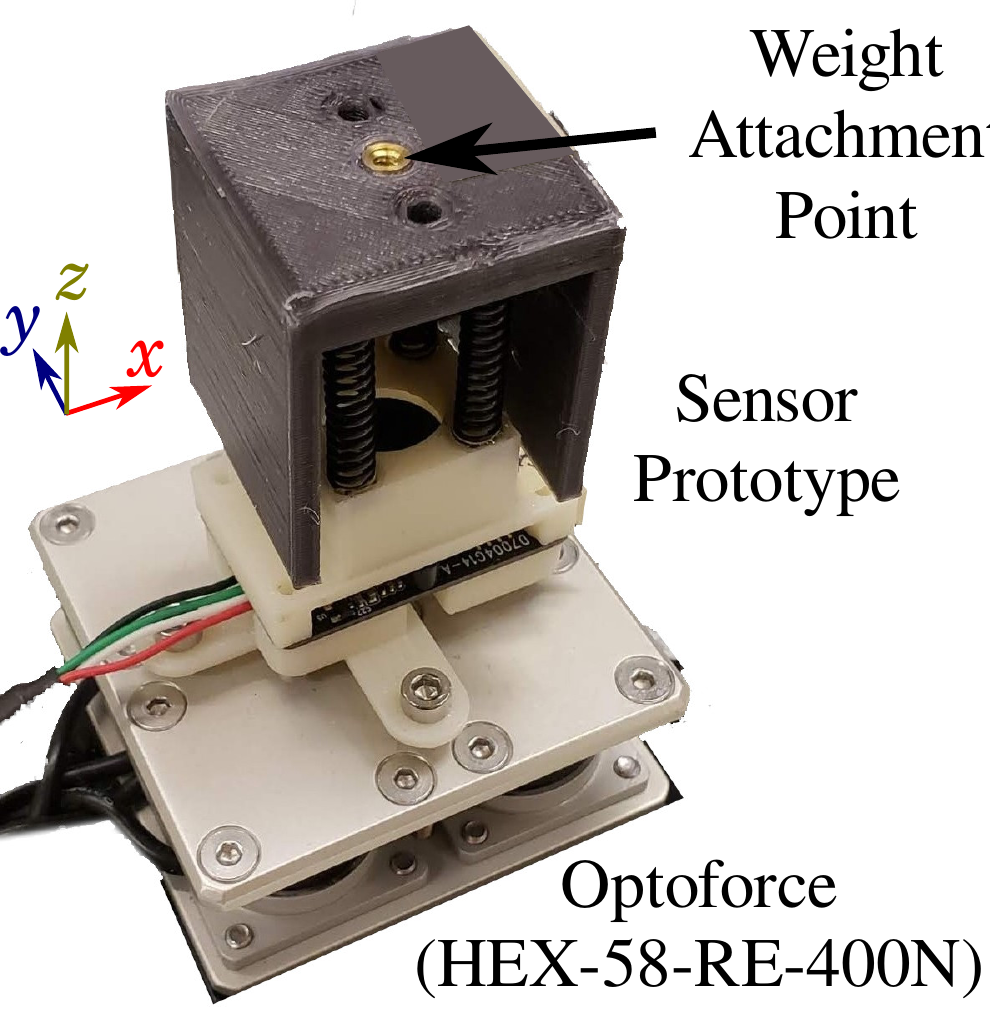}
            \caption{Experimental setup.}
        \end{subfigure}
        \hfill
        \begin{subfigure}[b]{0.45\linewidth}
            \centering
            \includegraphics[width=0.9\linewidth]{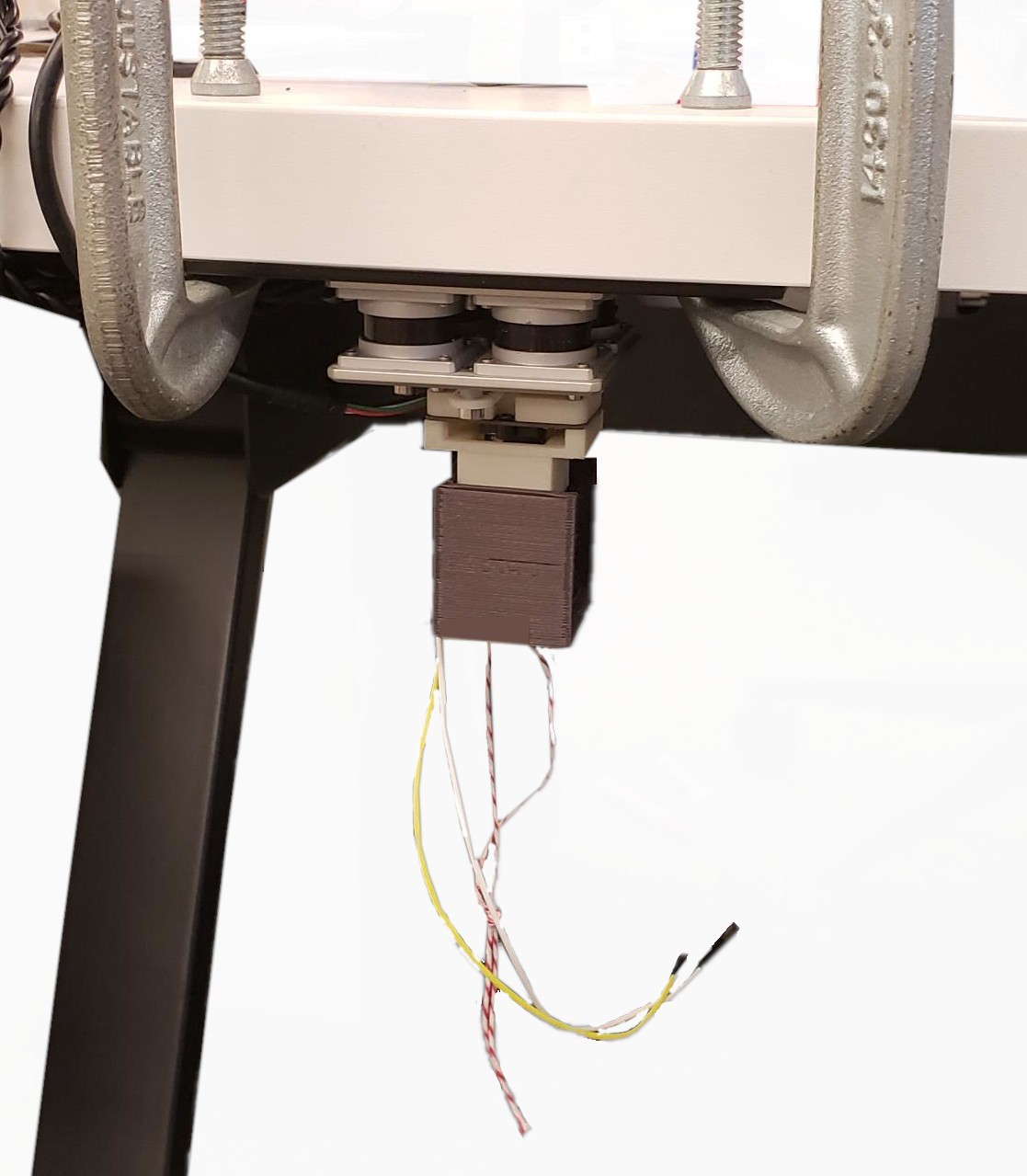}
            \caption{Calibration method.}
        \end{subfigure}
        \caption{Left, the data collection setup is shown (with the
            LED off -- note that out-of-frame, there is an Arduino supplying
            $\SI{3.3}{V}$ to the LED. Later designs used a 3.3~V coin cell
            battery to make the sensor standalone). Right, a method to
            calibrate the sensor without using the commercial sensor is
            demonstrated. The sensor is mounted upside down and weights are hung
            by string from the sensor to apply force uniaxially to the +$z$ axis.}
        \label{fig:datacollect}
    \end{figure}

    In order to evaluate the linearity (and therefore usefulness) of the sensor, we used a commercial
    force-torque sensor (Model HEX-58-RE-400N, OptoForce, Budapest, Hungary) to
    provide ground truth measurements. Although the OptoForce measures force and
    torque at a different origin than where the load is applied,
    the analysis of the linearity of the sensor holds. Data was collected with a
    Python script which used the OpenCV library to interface with the camera.
    The setup is shown in \cref{fig:datacollect}.

    Autocorrelation was used to determine the lag between our sensor and the
    OptoForce. The sensor lag between the prototype sensor and the OptoForce was
    roughly 40~milliseconds. Next, linear interpolation was used to match our
    sensor data with the OptoForce data, which were output at roughly 25~Hz and
    125~Hz respectively. The sensor data was smoothed with an
    exponential filter with weight of 0.2 to improve the autocorrelation results. 

    For calibration, we take a dataset of displacements $D$ and apply linear
    regression (with an affine term) against all six axes.
    $\theta$, $\phi$, and $\gamma$ refer to rotation around the $x$, $y$, and
    $z$ axes respectively. $K$ then forms a 6-by-6 matrix as shown below.
    \begin{align}
        \begin{bmatrix}
            F_x \\ F_y \\ F_z \\ M_x \\ M_y \\ M_z
        \end{bmatrix}
        =
        \left[
        \begin{array}{cccccc}
            \cr \\
            & & K_{6\times 6} & & & \\
            \cr \\
        \end{array}
        \right]
        \begin{bmatrix}
            D_x \\ D_y \\ D_z \\ D_{\theta} \\ D_{\phi} \\ D_{\gamma}
        \end{bmatrix}
        +
        \begin{bmatrix}
            \cr \\
            B\\
            \cr \\
        \end{bmatrix}
    \end{align}

    \subsection{Bandwidth}
    \label{sec:fps}

    Sensor bandwidth is directly limited by the camera framerate.  This must be
    physically measured since the Python script will output at unrealistically
    high framerate -- the OpenCV library reads from a buffer of stale images and
    will return a result even if the camera has not
    physically delivered a new frame.
    The webcam is pointed at a display with high refresh rate. A script turns
    the screen black, and as soon as the camera detects the black color, the
    screen changes to white, and so forth, and the frames displayed is compared
    to system time to obtain the framerate of the webcam. 

    Note that this calculates our maximum sensor bandwidth; our actual sensor
    bandwidth is determined by the tag detection rate. If dynamic
    instead of quasi-static loading is assumed, then motion blur can lead to tag
    detection failure.

    \section{Results}

    \subsection{Linearity}

\begin{figure}[thpb]
            \includegraphics[width=0.49\textwidth]{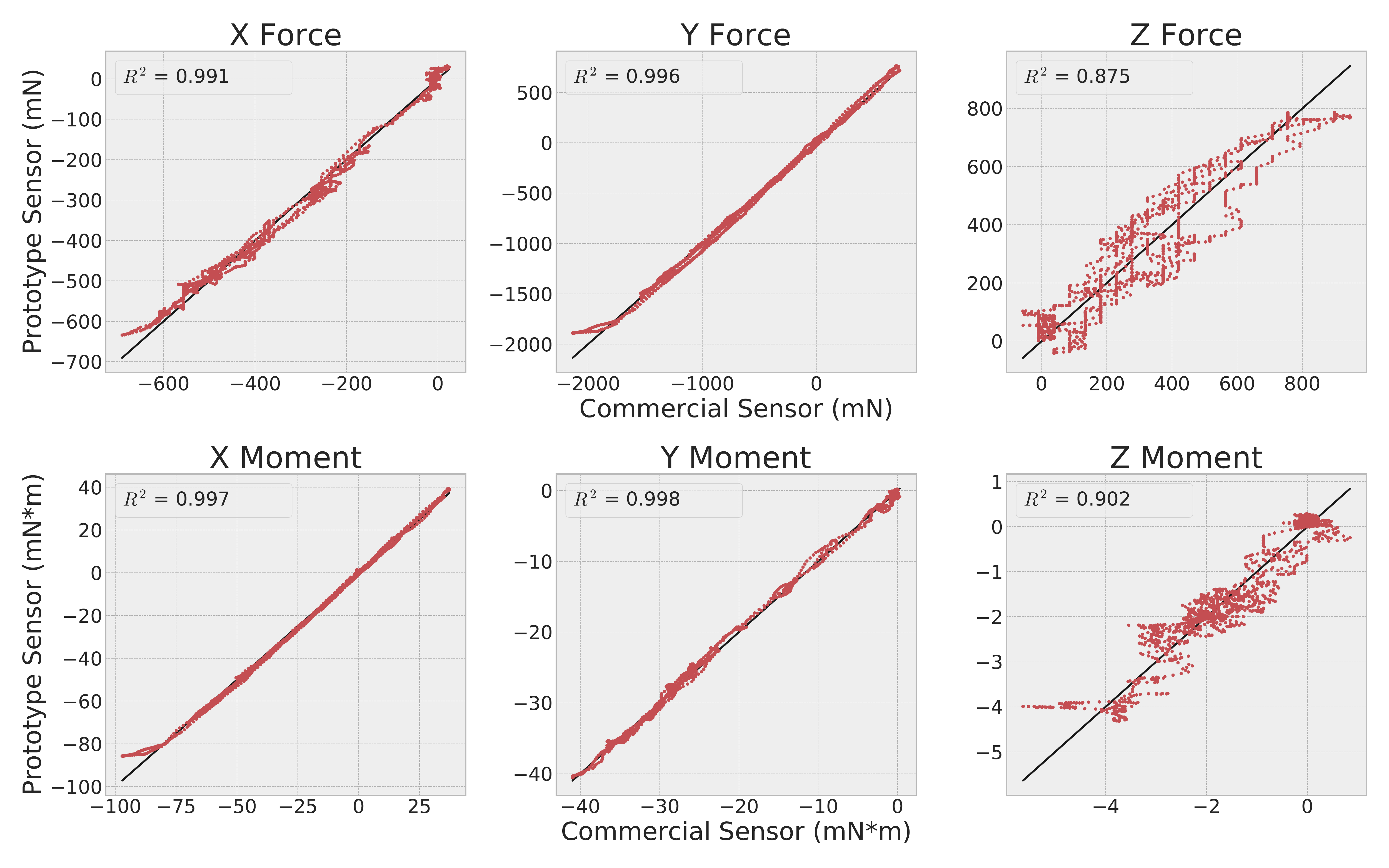}
            \caption{The black line represents a perfectly linear response between our
                sensor and the commercial sensor. The red dots show the actual
            sensor measurements using the ArUco tags.}
            \label{fig:linfit}
    \end{figure}

    In multiaxial loading, the sensor was manually moved around in all
    directions. As shown in \cref{fig:linfit}, the fits had a $R^2$ of 0.991,
    0.996, 0.875, 0.997, 0.997, and 0.902 for the $F_x, F_y, F_z, M_x, M_y$, and
    $M_z$ axes respectively.  The $F_z$ axis fit is notably worse than the $F_x$
    and $F_y$ fits, which was expected as explained in \cref{sec:zres}. 

    For qualitative comparison, \cref{fig:linfit}
    shows an example of a reconstructed dataset, where the linear fits are plotted
    against the original signal for qualitative comparison. This diagram shows
    the relatively large deviations in $F_z$ from the original signal, indicating
    noisiness in the tag measurements. 
    
    \begin{figure}[thpb]
            \includegraphics[width=0.49\textwidth]{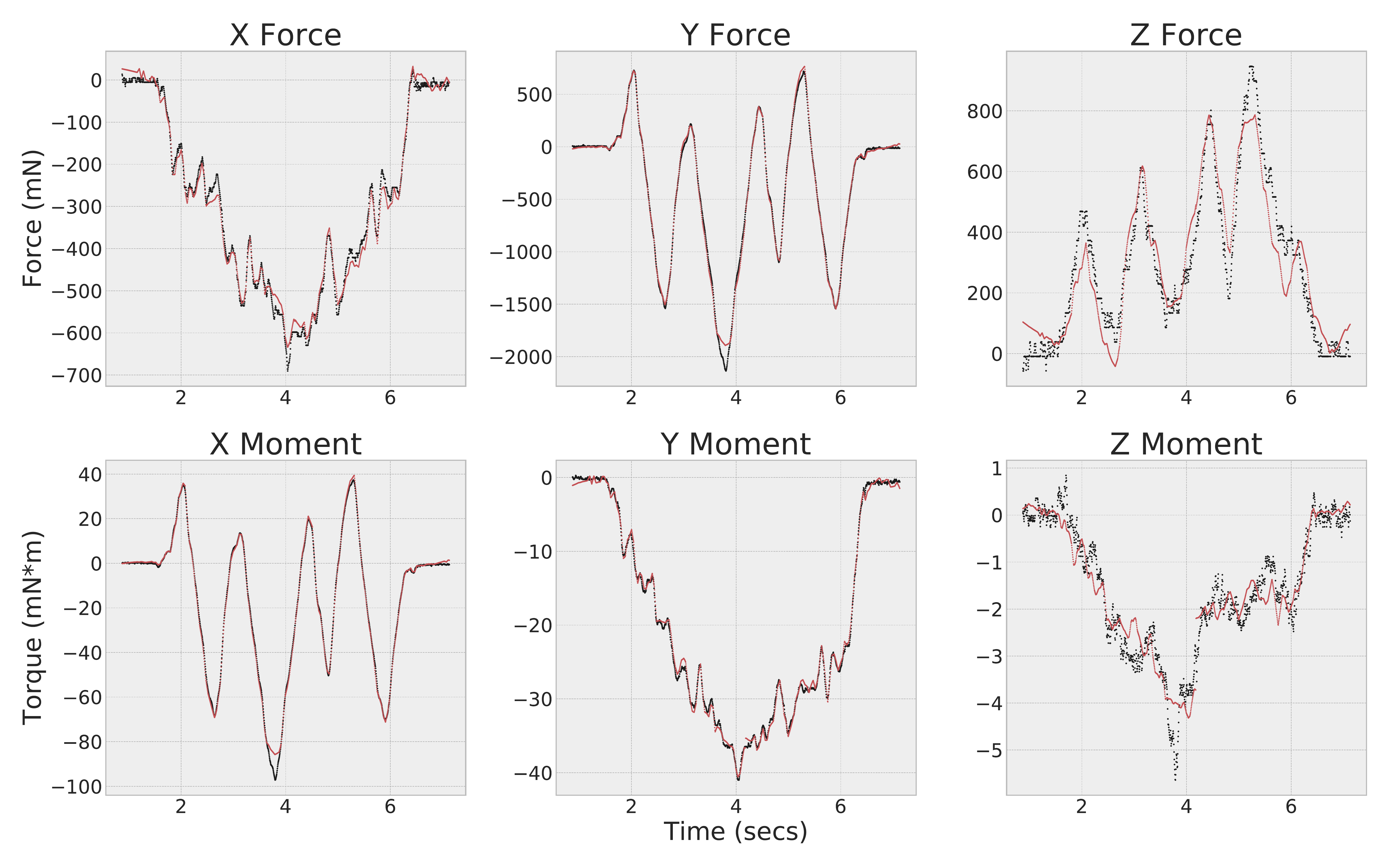}
            \caption{For qualitative inspection, a compound-loading dataset 
            is shown here. The commercial sensor measurements are in black, and
        the interpolated and linearly fitted prototype sensor's measurements are shown in red. } 
            \label{fig:sampledata}
    \end{figure}

    \subsection{Bandwidth}

    Our maximum sensor bandwidth is experimentally
    determined to be 25~Hz. Additionally, the camera we used was one of three cameras bought 
    by selecting for low cost, quick availability, and lack of external camera
    case. We also measured the other two cameras which, despite advertising
    similar framerates, exhibited noticeable differences in framerate. Operating
    at 640x480, we measured 25 fps, 33 fps, and 15 fps for the three cameras, as
    listed in \cref{tbl:camera}.     
    \begin{table}[htbp]
        \vspace{2mm} 
        \caption{Camera Specifications}
        \label{tbl:camera}
        \centering
            \resizebox{0.48\textwidth}{!}{
                \begin{tabular}{@{}llllll@{}}
                    \toprule
                    Camera Module Name    & Nominal Max Res. & Price & Year &
                    FPS @ 640p\\ \midrule
                    D7004G14-A (ours)      & 1280*720p@30fps  & \$20 & N/A  & 25\\
                    OV 2710         & 1920*1080@30fps  & \$20   & 2017 & 33\\
                    ELP Super Mini  & 1280*720p@30fps  & \$30   & 2015 & 15\\ \bottomrule
                \end{tabular}
            }
    \end{table}

    \section{Discussion}

    Our prototype sensor showed mostly linear responses
    under dynamic loading. While the linearity is not precise, these results
    still validate the underlying hypothesis that with fiducials it is possible
    to collect data on all three axes of force and three axes of torque. Further
    design iterations could improve on these results, although this approach is unlikely to 
    achieve the ~0.1\
    sensors.

    \subsection{Design Goals}
    \label{sec:conclusion}

    The sensor can now be evaluated against the goals specified previously in
    \cref{sec:designgoals}. The sensor design is indeed responsive in all six axes
    (after our pivot from one tag to two tags, as well as using a much brighter
    LED). Additionally, for grasping applications, the calculations in
    \cref{eqn:maxdisplacement} shows that if a much stiffer spring were chosen
    so that 40~N of load could be applied without exceeding the $y_{range}$,
    the sensor would still have better than 0.1~N of sensitivity.

    The qualitative design goals were also met. The sensor is small,
    measuring only 3.6~cm by 3.1~cm by 5.1~cm in size.
    The sensor is inexpensive, with the majority of the cost being a \$20
    webcam. The sensor is robust and has survived multiple plane trips and the
    occasional throw or drop. The sensor is also easy to modify. The light shield
    can easily be unbolted to
    change the fiducials, or re-printed in an hour to accommodate different
    designs (e.g. a single-tag vs. dual-tag design). Fabrication is easy and
    non-toxic, requiring no degassing machine (as with elastomer-based sensors)
    nor electrical discharging machines (as with custom strain-gauge based designs).
    The sensor by design does not suffer from thermal considerations (as in
    \cite{Guggenheim2017RobustAI}) or electrical noise (as with designs based on
    strain gauges). 

    \subsection{Error Sources}

    An important consideration is the coordinate origin around which measurements
    are made. As load must be applied to the spring platform on which the tags
    are glued, the origin around which measurements are collected may be
    different than desired, although a linear offset matrix should suffice to
    correct for this. Our six-axis measurement reflects a combination of a camera
    pose estimation and mechanical coupling, each of which can introduce errors.
    In the following section on sensor improvement, we focus on camera sensor issues.

    \subsection{Sensor Improvements}

    \subsubsection{Fiducial Changes}

    Unlike the standard use cases for ArUco markers, we do not care about
    distinguishing multiple objects and care more about the quality of the pose
    estimate for a tag guaranteed to be in-frame. A custom fiducial (perhaps
    solely a checkerboard) could improve the force-torque measurements. 

    \subsubsection{Noise in z-axis}

    The sensor is noisy in force and torque measurements along the $z$-axis.
    To address this, one possibility is to use a mirror and two tags which are laid flat
    on the $xy$ plane and the $yz$ plane respectively. The ``sideways" tag (on
    the $yz$ plane) has good sensitivity to $z$-axis displacements, and the flat
    $xy$ plane tag is addresses rotations around the $z$-axis. A 45-degree
    mirror then allows the camera to also observe the "sideways" tag on the $yz$
    plane. On the downside, the small mirror could make assembly difficult.

    \subsubsection{Sensor Size}

    Closer placement of the tag, to minimize the size of the sensor, may also be
    desired – this would necessitate a custom lens for the camera to allow for
    closer focus (e.g. a macro lens).  Miniaturization could also be
    accomplished with a smaller camera, as in \cite{WardCherrier2018TheTF}. 

    \subsubsection{Replacing Springs}

    The use of springs means that the sensor may behave poorly in
    high frequency domains. Replacing the springs with another mechanism, such
    as a Stewart platform, could allow custom tuning of the
    response.  Another possibility would be to fill
    the gap between the camera and the tag with optically clear material that
    would be resistant to high frequency inputs. \cite{Kiva2016FingernailWS}
    used a similar idea with a magnet and hall effect sensor, for a three-axis force
    sensor. However, such a design would complicate fabrication and potentially
    make camera calibration difficult due to image warping.

    \begin{table}[t]
        \vspace{2mm} 
        \caption{List of components and approximate costs.}
        \label{tbl:bom}
        \centering
            \resizebox{0.49\textwidth}{!}{
                \begin{tabular}{lll}
                    \toprule
                    Part          & Details   & Cost  \\ \midrule
                    Camera        & Mini Camera module, AmazonSIN: B07CHVYTGD& \$20 \\
                    LED and 2 wires & Golden DRAGON Plus White, 6000K, 124 lumens& \$2 \\
                    4 springs     & Assorted small springs set & \$5  \\
                    3D printed pieces  & PLA filament  & \$5  \\
                    Heat-set Threaded Inserts       & Package of 50 from McMaster-Carr (use 2) & \$1  \\
                    Misc. Bolts   & Hex socket head      & \$1  \\
                    Epoxy         & 5 minute             & \$5  \\ \bottomrule
                \end{tabular}
            }
    \end{table}

\section{Conclusion}
    We present a novel type of six-axis force-torque sensor using fiducial tags
    and a webcam. The design is fast to fabricate and simple to use, and is also
    strong enough to survive drops and crashes common in contact-rich tasks such
    as robotic grasping.  With only 3D-printed custom components, the design
    needs minimal technical expertise to adapt to applications ranging from
    manipulation to human-computer interaction research. The open-source design
    also allows for direct integration in designs for tasks such as grasping
    where sensor size is important.  This fiducial-based sensor is less accurate
    than commercial force-torque sensors, but is also orders-of-magnitude less
    expensive -- commercial sensors can cost thousands of dollars, while the
    parts cost of our sensor is under \$50 (see \cref{tbl:bom}). These combined
    advantages of our prototype sensor validates the general design principle of
    using 3D pose estimates from printed fiducials to create a six-axis
    force-torque sensor. Future work on improving the $F_z$ and $M_z$ axes could
    allow for an inexpensive, user-friendly, and robust alternative to current
    commercial sensors, opening up a new range of use cases for six-axis
    force-torque sensors.

\newpage
\bibliographystyle{IEEEtran}
\bibliography{IEEEabrv,references}
\end{document}